\def\year{2022}\relax
\documentclass[letterpaper]{article} 
\usepackage{aaai22}  
\usepackage{times}  
\usepackage{helvet}  
\usepackage{courier}  
\usepackage[hyphens]{url}  
\usepackage{graphicx} 
\usepackage{natbib}  
\usepackage{caption} 
\DeclareCaptionStyle{ruled}{labelfont=normalfont,labelsep=colon,strut=off} 
\usepackage{subdef}
\usepackage{multirow}
\usepackage{algorithm}
\usepackage{algorithmic}
\usepackage[dvipsnames]{xcolor}
\usepackage{subcaption}
\usepackage[frozencache=true,cachedir=.]{minted}

\definecolor{LightGray}{gray}{0.95}
\colorlet{ResultColor}{SpringGreen}
\setminted[prolog]
{
framesep=2mm,
bgcolor=LightGray,
fontsize=\scriptsize
}

\usepackage{newfloat}
\usepackage{listings}
\floatstyle{ruled}
\newfloat{listing}{tb}{lst}{}
\floatname{listing}{Listing}
\title{AUTO-DISCERN: Autonomous Driving Using Common Sense Reasoning}

\title{AUTO-DISCERN: Autonomous Driving Using Common Sense Reasoning}
\author {
   Suraj Kothawade,\textsuperscript{}
   Vinaya Khandelwal,\textsuperscript{}
   Kinjal Basu,\textsuperscript{}
   Huaduo Wang,\textsuperscript{}
   Gopal Gupta\textsuperscript{}
}
\affiliations {
    \textsuperscript{} Computer Science Department, The University of Texas at Dallas, Richardson, USA\\
    \{suraj.kothawade, vinaya.khandelwal, kinjal.basu, huaduo.wang, gupta\}@utdallas.edu
}

\newcommand{\model}{\mbox{\textsc{Auto-Discern }}}
\newcommand{\figref}[1]{Fig.~\ref{#1}}
\newcommand{\tabref}[1]{Tab.~\ref{#1}}
\newcommand{\secref}[1]{Sec.~\ref{#1}}

\begin{document}

\maketitle

\begin{abstract}
Driving an automobile involves the tasks of observing surroundings, then making a driving decision based on these observations (steer, brake,  coast, etc.). In autonomous driving, all these tasks have to be automated. Autonomous driving technology thus far has relied primarily on machine learning techniques. We argue that appropriate technology should be used for the appropriate task. That is, while machine learning technology is good for observing and automatically understanding the surroundings of an automobile, driving decisions are better automated via commonsense reasoning rather than machine learning. In this paper, we discuss (i) how commonsense reasoning can be automated using answer set programming (ASP) and the goal-directed s(CASP) ASP system, and (ii) develop the \model \footnote{\model : \textbf{AUTO}nomous \textbf{D}riv\textbf{I}ng u\textbf{S}ing \textbf{C}ommon s\textbf{E}nse Reasoni\textbf{N}g} system using this technology for automating decision-making in driving. The goal of our research, described in this paper, is to  develop an autonomous driving system that works by simulating the mind of a human driver. Since driving decisions are based on human-style reasoning, they are explainable, their ethics can be ensured, and they will always be correct, provided the system modeling and system inputs are correct. 
\end{abstract}
\section{Introduction}

\begin{figure}[t] 
    \centering
    \includegraphics[scale = 0.8]{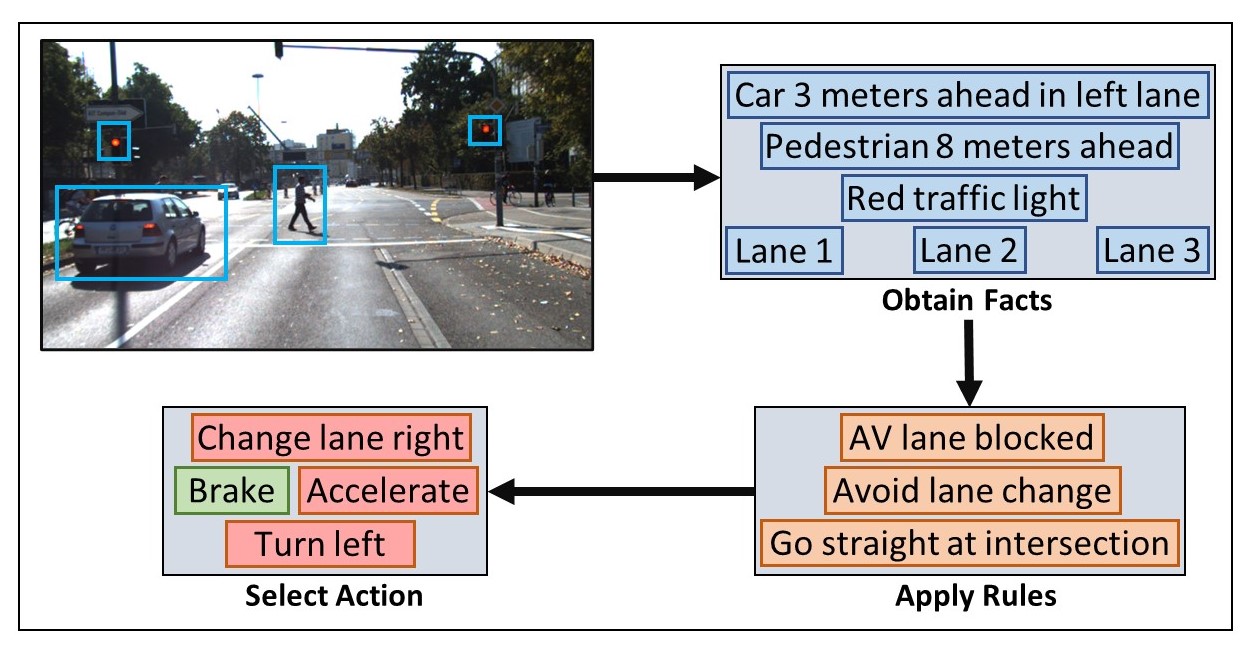}
    \caption{Overview of the \model system.}
    \label{model-overview}
    \label{fig:overview}
    \end{figure}

Autonomous Vehicles (AVs) have been sought for a long time. With the availability of cheaper hardware (sensors, cameras, LIDAR) and the advent of advanced software technology (AI, Machine/Deep learning (ML/DL), Computer Vision) over the last decades, rapid advancements have been made in AV technology. However, no car has yet achieved full automation or Society of Automation Engineers (SAE) Level 5 \cite{sae-av}. We believe that AV technology advancement has slowed due to over-reliance on ML/DL for automating all aspects of AVs. While ML technologies are important for developing AV technology, we believe that we can achieve better success by closely emulating how humans drive a car. Once a human driver has viewed their surrounding and processed a scene in their mind, they use their \textit{commonsense knowledge} and \textit{commonsense reasoning} to make driving decisions (e.g., if the traffic light is red, apply brakes and stop). Our goal in this paper is to develop an AV system that emulates the mind of a human: we will use ML/DL technology for tasks for which humans use pattern matching (vision and scene understanding) and automated commonsense reasoning for tasks for which humans perform mental reasoning (driving decision-making)(see \figref{fig:overview}).

To automate commonsense reasoning, we use ASP \cite{gelfond2014knowledge,cacm-asp,clingo} and the goal-driven implementation of ASP called s(CASP)  \cite{scasp}. A goal-driven implementation of ASP is important for automated commonsense reasoning as SAT-solver based implementations such as CLINGO \cite{clingo} face several practical issues (e.g., scalability, explainability) \cite{arcade} for applications such as autonomous driving. 

\section{Autonomous Vehicle Technology}

We express various decision making strategies that drivers use as commonsense rules in ASP, that will be executed on the s(CASP) system. These rules capture various driving decisions regarding steering, turning, braking, accelerating, stopping, etc.  We also report on a prototype system called \model that we have developed that takes a scene-description and sensor values as input and calculates the driving decision at that instant using the rules. A use case is shown in \figref{fig:overview}. We expect that a scene description (perception) will be obtained via image processing techniques that use deep learning methods. Because our decision making is based on automated commonsense reasoning, every decision can be explained, and, in theory, our system will never make a wrong decision, as long as the rules are correct and system input is correct. 

Our main contribution in this paper is to show how automated commonsense reasoning can be harnessed to achieve an SAE Level 5 autonomous driving system. 

Autonomous vehicles (AVs) hold enormous promise: they can reduce the cost of transportation and increase convenience, as well as reduce road accidents and traffic fatalities by a significant number. AVs can also have a big environmental impact: private ownership of cars can become unnecessary. AVs can greatly aid in the mobility of elderly, disabled, and disadvantaged. The story of AVs has been one of great optimism: it was projected in 2017 that there will be 10 million AVs on the road by 2020 and that fully autonomous vehicles will be the norm in 10 years. 
Not only we have not reached the 10M vehicle mark, but also no vehicle has earned the fully automated (SAE Level 5) designation.

The history of making vehicles autonomous began with the introduction of the cruise control in 1948 \cite{history-av}. In 1999, the US Federal Communication Commission allocated 75 MHz of spectrum dedicated to short range communication. In early 2000s, several teams developed and demonstrated autonomous cars in response to a DARPA grand challenge \cite{DARPA-grand-challenge1,DARPA-grand-challenge2}. In 2009, Google began its self-driving project, and in 2014 Google's AV passed a 14-mile driving test in Nevada. 
The US National Highway Transportation and Safety Administration (NHTSA) released its initial policy on AVs that year, and in 2015, Tesla released its autopilot self-driving software. Since then, many other companies have entered the market, and many partnerships have been forged between them. The Society of Automation Engineers developed a scale for vehicle automation: 
    Level 0: No automation; 
    Level 1: Driver assistance (e.g., cruise control);
    Level 2: Partial automation (perform steering and acceleration under human watch);
    Level 3: Conditional automation (most tasks can be performed, but human driver has control;
     Level 4: High automation (car can perform all tasks under specific circumstances, e.g., in a geo-fenced area; human over-ride is still possible);
    Level 5: Full automation (all tasks automated; no human intervention required at all).

\noindent No car has reached SAE Level 5 yet. The principle reason, we believe, is over-reliance on ML and DL technologies for most aspects of driving. Our goal here is to show that automated commonsense reasoning is essential for achieving SAE Level 5 automation.


\section{Machine Learning-based AV Systems}

At present, machine learning plays a major role in the AV technology. A significant amount of technology goes into an AV: radar (to detect cars and other large objects), ultrasonic sensors (to detect objects close by, e.g, the curb), lidar (to detect lane markings, edge of a road, etc.), GPS (for direction to destination and knowing AV's location), and video cameras (for obtaining the surrounding scene and analyzing it). All these are connected to a central computer where processing takes place. Driving data is collected along with sensor readings, video images, Lidar data, etc. In NVidia's PilotNet project \cite{bojarski2020nvidia}, for example, a deep learning model is trained on this data to make one of three decisions: steering along a predicted trajectory, amount of braking, and amount of acceleration. 
Predicted trajectories are one of the following: (i) lane stable (keep driving in the same lane); (ii) change to left lane (first half of left-lane-change maneuver); (iii) change to left lane (second half of left-lane-change maneuver); (iv) change to right lane (first half of right-lane-change maneuver); (v) change to right lane (second half of right-lane-change maneuver); (vi) split right (e.g. take an exit ramp); (vii) split left (e.g., left branch of a fork in the road). A ML based solution can be boiled down to predicting the degree of steering, the amount of braking, and the amount of acceleration at every moment during driving. Of course, the last two are mutually exclusive, that acceleration and braking are rarely needed at the same time.

\begin{figure}
    \includegraphics[scale = 0.425]{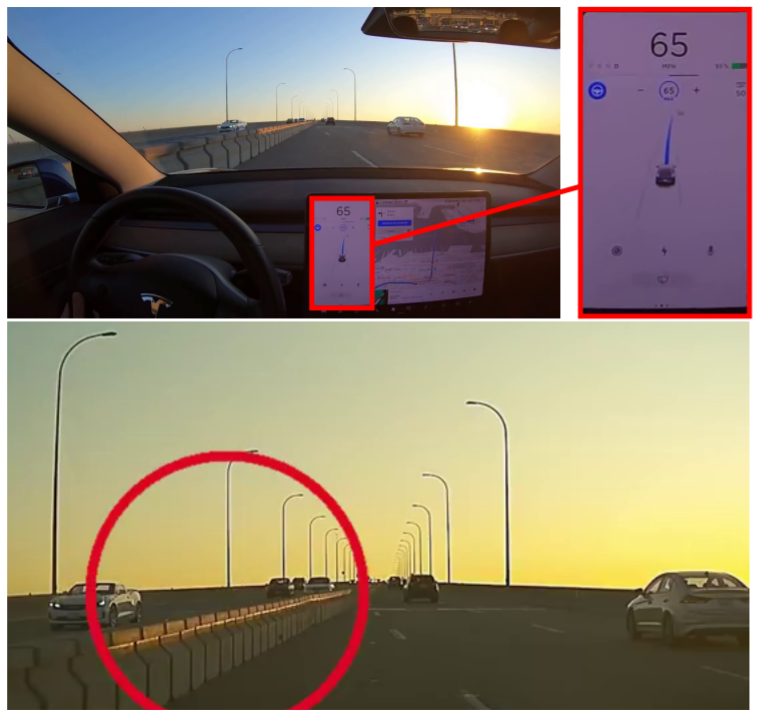}
    \caption{Tesla self-driving fails to perform a lane merge right. The radar sensor detects a barrier to the left. However, the visual component is confused by the reflection on the barrier (likely thinks of it as a yellow dividing lane marking) and does not register that the lane is ending.}
    \label{fig:tesla-av-confused}
    \vspace{-2ex}
\end{figure}
    
\begin{figure}
    \captionsetup[subfigure]{labelformat=empty}
    \centering
    \begin{subfigure}[b]{0.47\textwidth}
        \includegraphics[width=0.5\textwidth]{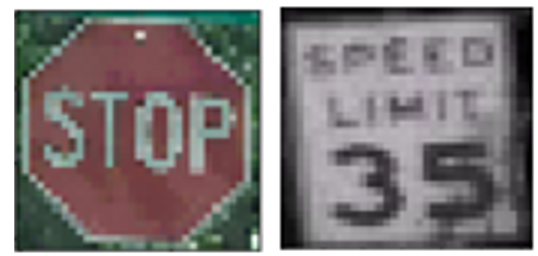}
        \hfill
        \includegraphics[width=1.9cm, height=1.9cm]{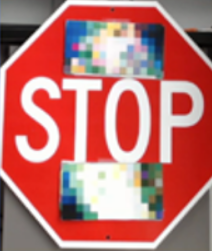}
        \hfill
        \includegraphics[width=1.9cm, height=1.9cm]{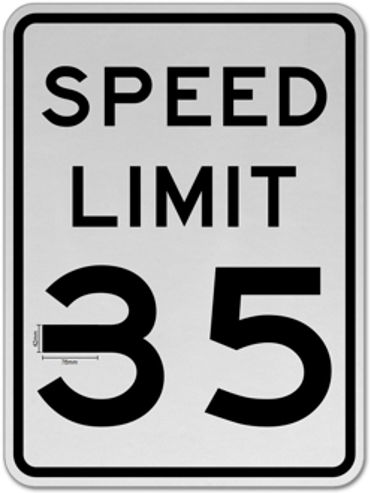}
    \end{subfigure}
    \caption{Partial modifications to traffic signs can cause ML models to misclassify the sign entirely. A stop sign can be classified into a speed limit. Speed limits can be misinterpreted as well, a small tape can make 35 to be classified as 85.}
    \label{fig:modified-signs}
    \vspace{-2ex}
\end{figure}

Neural-based technology has a number of well-known issues with respect to accurate prediction of an outcome. An ML algorithm is a universal function that learns an approximate mapping from an input to an output in accordance to the training data. This technology is fundamentally statistical. Edge cases, unusual circumstances that are uncommon in training data, are not covered. Obviously, anything outside the training data is not learned. This is especially true for autonomous driving, where there are many situations that may never be encountered in the driving data collected for training. A good example of this is flashing red and blue lights in police cars and fire trucks that Tesla autopilot system has had trouble with \cite{tesla-police-firetruck}.
There are many such examples where a ML system gets  confused by the noise present in the data and generates an erroneous model. \figref{fig:tesla-av-confused} shows a scenario where Tesla AV model fails to detect a barrier on the left. \figref{fig:modified-signs} shows an example where slight perturbations to traffic signs can cause failure cases in ML models. There are many instances where an AV got into an accident, resulting in loss of life \cite{dangers}. We believe that our commonsense reasoning-based \model system can safely deal with situations where other ML-based systems failed (see error mitigation in \secref{sec:discussion}).


Because of these issues, many companies have scaled down their ambitions down from SAE level 5. Some companies shut down (e.g., Starsky Robotics). Others (e.g., Waymo) revised their goal down to achieving SAE Level 4 while others have restricted  autonomous behavior within limited circumstances (e.g., geo-fenced areas). 
So the goal of reaching SAE Level 5 seems illusive. In this paper, we argue that SAE Level 5 can be reached via automated commonsense reasoning. In fact, we believe that automated commonsense reasoning is indispensable for the AV technology to reach SAE Level 5 (full automation).


\section{AV based on Commonsense Reasoning: Motivation}  

To drive a vehicle, a human driver must have the:

\begin{enumerate}
    \item ability to control the vehicle, i.e., be able to steer, brake, accelerate, and signal for a turn.
    \item ability to make visual deductive inferences, i.e., be able to see objects in front of or around the vehicle and make decisions, estimate the speed of objects, and project where they will be in the near future.
    \item ability to make a visual abductive inference, i.e.: (i) be able to infer hidden or occluded parts of the objects; e.g., a car will normally have four tires, even though only two are visible; (ii) be able to perform counterfactual (``what-if”) reasoning.
    \item ability to distinguish between various scenarios, e.g., be able to tell, for example, that a car in a bill board is not the same as a car on the street.
\end{enumerate}


Essentially, there are two types of tasks involved in driving:  (i) visual scene processing and inferencing (tasks \#2 through \#4 above) and (ii) controlling the vehicle (task \#1). \textit{Learning to control the vehicle is hard for humans, but visual inferencing comes naturally to us. Controlling the vehicle is easy for a machine, learning visual inferencing is significantly harder for it.}

To realize truly autonomous vehicles (SAE Level 5), \textit{we need to emulate the way humans drive cars}, i.e., use ML for scene processing, while using automated commonsense reasoning for making inferences regarding driving actions. Once an inference is made (steer, accelerate, brake, etc.), it can be easily carried out by the machine. Our ideas are based on the insight that for tasks for which humans use pattern matching (e.g., picture recognition), we should use ML technology, while for tasks for which humans use deduction, we should use automated reasoning. The current practice of using ML for all AV tasks is overkill and is preventing us from reaching SAE Level 5.

We envisage that the cameras will provide an image of the surroundings every second or so. This image will be processed using deep learning (object, lane detection, depth prediction, etc.) so that all the items present in the picture will be labeled and their bounding-boxes marked. The labels will be predicates, that will be extracted from the picture, along with coordinates of each bounding-box. A picture can be labeled with predicates with the help of datasets such as Visual Genome \cite{genome} and systems such as DenseCap \cite{densecap}. These predicates that describe the picture, constitute the input to the \model system.
The commonsense rules that a human driver uses will take this data (expressed as predicates that capture the position and spatial relationship among various objects in the scene) as input to compute a driving decision.  The question then arises: \emph{How does one automate commonsense reasoning?}

\section{Background}


\textbf{Commonsense Reasoning:}
As mentioned earlier, an autonomous driving system should be able to understand and reason like a human driver. If we examine how we humans reason, we fill lot of gaps in our understanding of a scene, conversation, or a piece of text we read, through our commonsense knowledge and reasoning (e.g., if we see a car moving fast on a road, we use our commonsense knowledge to infer that, normally, there must be a driver inside). Thus, to develop autonomous driving software, we need to automate commonsense reasoning, i.e., automate the human thought process. The human thought process is flexible and \textit{non-monotonic} in nature, which means \textit{ ``what we believe now may become false in the future with new knowledge''}. It is well known that commonsense reasoning can be modeled with (i) defaults, (ii) exceptions to defaults, (iii) preferences over multiple defaults, and (iv) modeling \textit{multiple worlds} \cite{gelfond2014knowledge,cacm-asp}. 

Much of human knowledge consists of default rules, for example, \textit{``Normally, birds fly''} is a default rule. However, there are exceptions to defaults, for example, \textit{penguins are exceptional birds that do not fly}. Reasoning with default rules is non-monotonic, as a conclusion drawn using a default rule may have to be withdrawn if more knowledge becomes available and the exceptional case applies. For example, if we are told that Tweety is a bird, we will conclude it flies. Knowing later that Tweety is a penguin will cause us to withdraw our earlier conclusion. Similarly, if we see a car, we know there must be a driver inside, normally, unless we realize it's a robot-taxi, then we withdraw that conclusion.

Humans often make inferences in the absence of complete information. Such an inference may be revised later as more information becomes available. This human-style reasoning is elegantly captured by default rules and exceptions. Preferences are needed when there are multiple default rules, in which case additional information gleaned from the context is used to resolve which rule to apply. One could argue that expert knowledge amounts to learning defaults, exceptions, and preferences in the field that a person is an expert in.

Also, humans can naturally deal with \textit{multiple worlds}. These worlds may be consistent with each other in some parts, but inconsistent in other parts. For example, animals don't talk like humans in the real world, however, in the cartoon world, animals do talk like humans. So Nemo the fish, may be able to swim in both the real world and the cartoon world, but it can talk only in the cartoon world. Similarly, we are able to distinguish between a car shown on a billboard on the road and an actual car on the road. Humans have no trouble distinguishing between multiple worlds (world shown in the billboard vs real world) and can easily switch between them as the situation demands. Default reasoning, augmented with the ability to operate in multiple worlds, allows one to closely represent the human thought process. Default rules with exceptions and preferences and multiple worlds can be elegantly realized in the paradigm of ASP \cite{gelfond2014knowledge,baral,cacm-asp} and executed using the s(CASP) system \cite{scasp}. 

\smallskip
\noindent \textbf{ASP and s(CASP):}
ASP is a declarative paradigm that extends logic programming with negation-as-failure. 
We assume that the reader is familiar with ASP.
Considerable research has been done on ASP since the inception in the late 80s of the stable model semantics that underlies it~\cite{cacm-asp}. A major problem with ASP implementations is that programs have to be grounded and SAT-solver-based implementations such as CLINGO~\cite{clingo} used to execute the propositionalized program to find the answer sets. There are multiple problems with this SAT-based implementation approach, which include exponential blowup in program size, having to compute the entire model, and not being able to  produce a justification for a conclusion~\cite{arcade}.\looseness-1

Goal-directed implementations of ASP such as s(CASP) ~\cite{scasp} work directly on predicate ASP programs (i.e., no grounding is needed) and are query-driven (similar to Prolog). The s(CASP) system only explores the parts of the knowledge-base that are needed to answer the query, and they provide a proof tree that serves as justification for the query. The s(CASP) system support predicates with arbitrary terms as arguments as well as constructive negation \cite{scasp}. It also supports abductive reasoning. Goal-directed implementations of ASP such as s(CASP) have been used for developing systems that emulate an expert. Chen et al have used it to emulate a cardiologists mind by automating applications of the guidelines they use for treating congestive heart failure \cite{chef}. The system, reportedly, can outperform cardiologists \cite{chef-ieee}. In our project, we want to emulate the mind of an automobile driver.

\begin{figure}
    \centering
    \includegraphics[scale = 0.5]{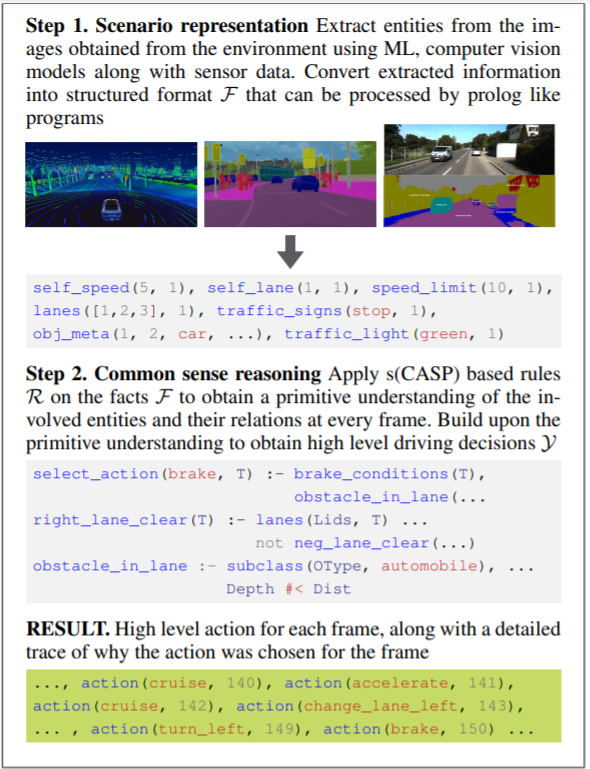}
    \caption{Computation Steps of \model.}
    \label{fig:computation-steps}
    \vspace{-2ex}
\end{figure}
    
\section{Commonsense Reasoning-based AV}    

The commonsense rules that a human driver uses for driving are modeled in ASP, using defaults, exceptions and preferences. The input to these rules is gleaned from a scene that the driver sees, where the scene is translated into a set of predicates that describe the objects and their placement in a scene. We assume that state-of-the-art ML technology is used to obtain these predicate labels. 
These predicates are represented as a set of facts in ASP that describe the environment and serves as input to the \model system. Formally, facts $\Fcal$ for a given frame at a timestamp $\Tcal$ are combined with (commonsense) driving rules $\Rcal$ to make a driving decision $\Ycal$, given an intent $\Xcal$. Fig. \figref{fig:computation-steps} shows the facts describing a scene and rule snippets that will be activated for this scenario to compute a driving decision at each timestamp.
The ASP facts $\Fcal$ contain speed, lane, relative distance, predicted trajectory of the AV and other detected objects, lane structure, intersection information, visible traffic signs and lights, etc.
An intent $\Xcal$ describes the short term goal that needs to be achieved by the AV in order to reach its destination. It is based on the instruction from the navigation system, for example, continue in lane, stay in leftmost lane, turn left, etc.
Based on the facts and the driving rules, a decision $\Ycal$ is taken which is one of accelerate, brake, cruise, change lane left, change lane right, turn left, turn right, etc.

\subsection{Driving Rules}
The driving rules written in s(CASP) are rules that drivers use while driving (e.g., if behind a slow-moving vehicle, change lanes to go faster [default], unless lanes are blocked [exception]). We refer to this collection of rules as the `Driving Rules Catalog'. First, we describe this catalog using examples. We then show (in the experimentation and testing section) how the catalog examples can be converted into ASP rules and executed in s(CASP) to make decisions. 

At the topmost level, the catalog is categorized by the set of actions (brake, accelerate, change lane, turn etc.). For each action, ASP rules have been developed that use knowledge of the scene to compute the driving action. Note that the number of commonsense rules that humans use while driving is not really that large. We have cataloged 35 rules at present. 
Next, we give some examples.

\begin{enumerate}
    \item[192] \textit{Change lane to left,} if there is a non-automobile obstacle ahead in the lane within x meters, and the left lane is clear to perform the lane change
    \item[193] \textit{Turn right,} if the intent is to enter the right lane, and if on the major lane of a T-junction intersection, and if AV's predicted path does not intersect with any object(pedestrian, cyclist, \ldots)
    \item[194] \textit{Turn right,} if the intent is to enter the right lane, and if at a signalized 4-way intersection, and if the traffic light is green, and if AV's predicted path does not intersect with any object
    \item[195] \textit{Brake,} if there is an object ahead, within stopping distance from the AV and in the same lane
    \item[196] \textit{Brake,} if at an unsignalized 4-way intersection, and AV vehicle is not the earliest to arrive at the intersection
\end{enumerate}

To arrive at a decision, the AV evaluates all possible actions consistent with the current intent and decides on the best one. 
For each action, there is a set of default rules and exceptions. The default rules evaluate the conditions under which AV should consider taking the action. The exception acts as a filter, checking if the action is safe to perform.
\\
The hierarchical logic and default reasoning have been encoded in s(CASP). Here is code for change lane-left: 

\begin{minted}{prolog}
select_action(change_lane_left, T) :-
    change_lane_left_conditions(T),
    not ab(d_select_action(change_lane_left, T)),
    not neg_select_action(change_lane_left, T).
    
select_action(accelerate, T) :- acc_conditions(T),
    not neg_select_action(accelerate, T).
select_action(change_lane_left, T) :- 
    change_lane_left_conditions(T),
    not neg_select_action(change_lane_left, T).
select_action(change_lane_right, T) :-
    change_lane_right_conditions(T),
    not neg_select_action(change_lane_right, T).
select_action(turn_left, T) :- turn_left_conditions(T),
    not neg_select_action(turn_left, T).
select_action(turn_right, T) :- turn_right_conditions(T),
    not neg_select_action(turn_right, T).
\end{minted}
The change lane left conditions define the default rules to perform the action. The AV would consider changing to left lane if the current intent is to stay in leftmost lane (precursor to performing a left turn) or it needs to overtake a vehicle or avoid an obstacle ahead.

\begin{minted}{prolog}
change_lane_left_conditions(T) :- self_lane(SLid, T),
    nonmv_ahead_in_lane(T, SLid, 20, OType),
    neg_can_drive_over(OType), can_swerve_around(OType).
change_lane_left_conditions(T) :- 
    intent(stay_in_leftmost_lane, T).
\end{minted}
However, it is not always possible to perform an action even if it is a short term goal. Predicate neg\_select\_action encodes the exception rules when performing the action is not safe or possible. The AV cannot change to the left lane if its not clear of obstacles. Similarly, it cannot accelerate if it is approaching a red traffic light or it is above the speed limit.

\begin{minted}{prolog}
neg_select_action(accelerate, T) :-
    above_speed_limit(T);
    self_lane(SLid, T), neg_lane_clear(T, SLid, 10);
    traffic_light(red, T).
neg_select_action(change_lane_left, T) :-
    not left_lane_clear(T).
\end{minted}















\begin{figure}[b]
   \centering
    \begin{subfigure}[b]{0.143\textwidth}
         \centering
         \includegraphics[width=\textwidth]{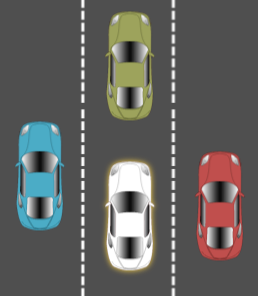}
         \caption{}
    \end{subfigure}
    \hfill
    \begin{subfigure}[b]{0.16\textwidth}
         \centering
         \includegraphics[width=\textwidth]{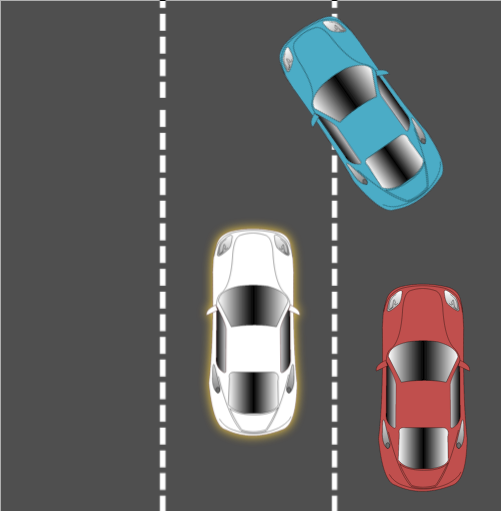}
         \caption{}
    \end{subfigure}
    \hfill
    \begin{subfigure}[b]{0.147\textwidth}
         \centering
         \includegraphics[width=\textwidth]{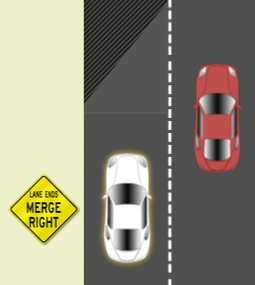}
         \caption{}
    \end{subfigure}
    \caption{Testing Scenario: white vehicle represents our AV (a) when obstructed by vehicles on all sides, AV should brake (b) when another object suddenly enters lane, AV should change lane if possible or brake (c) on a lane merge right, the AV should slow down, give way to traffic on target lane before merging}
    \label{fig:synthetic-scenarios}
    \hfill
\end{figure}

\subsection{Experimentation \& Testing}
We demonstrate the usability and simplicity of our approach to autonomous driving. Further, we show selected scenarios where common sense reasoning is essential to safe driving and how our approach achieves this. For each selected scenario, we show the relevant rules that lead to the action and describe the decision making process. 

\subsubsection{System Testing}
We have developed a large set of test scenarios to test our ASP-coded commonsense rules for driving. The test scenarios cover normal and adverse conditions encountered during driving. These scenarios cover general cases as well as corner case situations that even humans would find challenging to drive in. In \figref{fig:synthetic-scenarios}, we show some of the scenarios that are covered by our model.



\subsubsection{Common Driving Scenarios}

Once our rules were tested, we evaluated the \model system on real-world situations obtained from the \textsc{Kitti} \cite{Geiger2012CVPR} dataset. The experiment covers scenarios from a variety of traffic conditions that one may come across on a day-to-day basis. We test \model on manually annotated subset of \textsc{Kitti} scenarios to obtain a runtime analysis. We selected 3 representative frames from 2 scenarios per environment and show the runtime for each frame. \tabref{table:kitti-runtime} summarizes the result. The performance is largely determined by the number of objects in the frame and the complexity of the action performed.\footnote{Experiment run on Intel(R) Core(TM) i7-6500U CPU @ 2.50GHz 8-Core Processor, 12GB RAM.} Note that on average our system computes a decision for a frame in half a second. Our system will take a snapshot once per second. We expect, on average, half of this one second to go into analyzing the picture, annotating it and generating input data, and the other half in computing the driving decision using s(CASP).

\begin{table}[ht!]
\centering
\begin{tabular}{|c|c|c|}
\hline
\multirow{2}{*}{\shortstack{\textsc{Kitti}\\ Environment}} & \multicolumn{2}{c|}{\begin{tabular}[c]{@{}c@{}}Runtime\\ per frame (ms)\end{tabular}} \\ \cline{2-3} 
                             & \multicolumn{1}{c|}{Avg}               & \multicolumn{1}{c|}{Max}              \\ \hline
City                         & 413                                    & 873                                   \\
Road                         & 285                                    & 635                                   \\
Residential                  & 127                                    & 657                                   \\
Campus                       & 106                                    & 469                                   \\ \hline
\end{tabular}

\vspace{1em}

\begin{tabular}{|c|c|c|c|}
\hline
\multirow{2}{*}{\begin{tabular}[c]{@{}c@{}}\textsc{Kitti}\\ Scenario\end{tabular}} & \multicolumn{3}{c|}{Runtime (ms)} \\ \cline{2-4} 
                                                                          & Frame 1   & Frame 2   & Frame 3   \\ \hline
City 1                                                                    & 873       & 426       & 15        \\
City 2                                                                    & 507       & 525       & 262       \\
Road 1                                                                    & 160       & 26        & 32        \\
Road 2                                                                    & 635       & 631       & 342       \\
Residential 1                                                             & 657       & 21        & 16        \\
Residential 2                                                             & 25        & 24        & 159       \\
Campus 1                                                                  & 47        & 51        & 31        \\
Campus 2                                                                  & 469       & 84        & 16        \\ \hline
\end{tabular}

\caption{Run-times for \model on real-world environments in the \textsc{Kitti} dataset. Top table shows avg time taken for processing frames in various environments.
}
\label{table:kitti-runtime}
\end{table}

Selected frames from the experiment along with the rules that allowed \model to perform the required action are shown below. Fig. \ref{fig:kitti-1} shows a situation where the AV has to merge into the left lane with high incoming traffic. As the AV approached the merge, it slowed down to a stop, performing the lane change when the left lane was clear. Fig. \ref{fig:kitti-2} is an example of waiting before performing a right turn. The rules build upon predicted object trajectories obtained from ML model to realize performing a right turn is unsafe. These rules were derived from example 192 and 193 of the driving rules catalog respectively.

\begin{figure}[h]
    \centering
    \begin{subfigure}[b]{0.47\textwidth}
        \includegraphics[width=\textwidth]{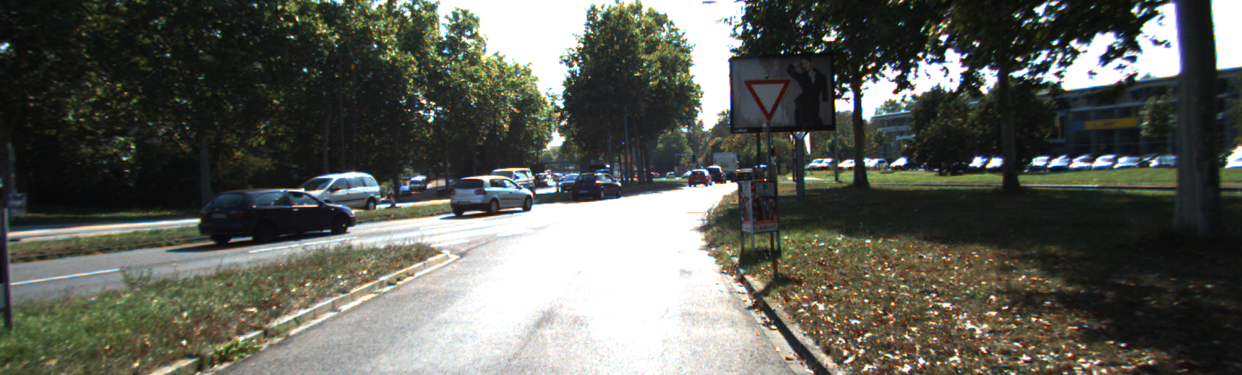}
        \vspace{-18pt}
        \begin{minted}{prolog}
change_lane_left_conditions(T) :-
    intent(merge_into_left_lane, T).
neg_select_action(change_lane_left, T) :-
    not left_lane_clear(T).
brake_conditions(T) :- intent(merge_into_left_lane, T),
    not left_lane_clear(T).
        \end{minted}
        \vspace{-18pt}
    \caption{AV performs a change lane left to merge at high city traffic.}
    \label{fig:kitti-1}
    \vspace{10pt}
    \end{subfigure}
    \begin{subfigure}[b]{0.47\textwidth}
        \includegraphics[width=\textwidth]{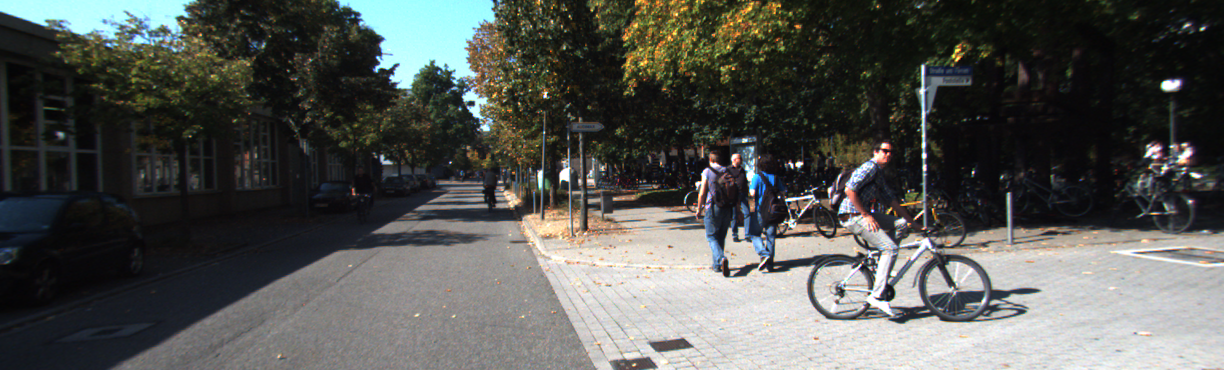}
        \vspace{-18pt}
        \begin{minted}{prolog}
turn_right_conditions(T) :- intent(enter_right_lane, T).
neg_select_action(turn_right, T) :-
    self_pred_path(SPath, T),
    obj_pred_path(Oid, OPath, T),
    path_intersects(SPath, OPath).
brake_conditions(T) :- intent(enter_right_lane, T),
    intersection(_, _, at, T).
        \end{minted}
        \vspace{-18pt}
        \caption{AV waits for pedestrians before performing a right turn.}
        \label{fig:kitti-2}
    \end{subfigure}
    \caption{Example frames from \textsc{Kitti} experiments.}
    \vspace{-4ex}
\end{figure}

Finally, we tested the \model system on cases where a ML based system failed. \model system was able to arrive at a correct decision in all such scenarios.




\subsection{Discussion} \label{sec:discussion}


Our experiments indicate that commonsense knowledge about driving can be modeled with relative ease. This should not come as a surprise, as we use relatively few rules to chart a course through the various types of objects we encounter as we drive. If the rules are correct, and the input to our system is correct, then we are in effect modeling an unerring human driver.
There are many advantages to developing an AV system based on commonsense reasoning:

\medskip\noindent{\bf Explainability:} Every driving decision made is explainable, as it can be justified via rules. The justification is obtained by using the s(CASP) system's proof tree generation facility \cite{arias-just}. An example proof tree fragment for the scenario in \figref{fig:kitti-1} is shown below:

{\small 
\begin{minted}[
    framesep=2mm,
    bgcolor=LightGray,
    fontsize=\scriptsize
]{text}
QUERY:Does 'start_drive' holds (for 410, and  411)?
>'start_drive' holds (for 410, and 411) because
 >'suggest_action' holds(for change_lane_left) because
  >'action' holds (for change_lane_left) and
   >there is no evidence that 'neg_suggest_action'
    holds (for change_lane_left) and
   >'select_action' holds (for change_lane_left) because
    >'change_lane_left_conditions' holds because
     >'intent' holds (for merge_into_left_lane).
    >there is no evidence that 'neg_select_action'
     holds (for change_lane_left) because
     >'left_lane_clear' holds because
      ...
      >there is no evidence that 'neg_lane_clear'
       holds (for Lid 2, and StopDist 10) because
      >there is no evidence that 'class' holds,
       with Var0 not equal bicycle, bike, car,
       pedestrian
       ...
The global constraints hold.
\end{minted}
}

\medskip\noindent{\bf Error Mitigation:} A major problem with ML-based systems is that unusual and corner cases may be missed. For example, as shown earlier, the speed limit of 35 may be read as 85. If a human misinterprets the speed limit sign of 35 as 85, their commonsense knowledge that they are in a city tells them that an 85 speed limit seems too high. This type of commonsense reasoning can be performed in our \model system. Defaults rules with exceptions can be written in ASP to determine what a reasonable speed limit ought to be in each type of surrounding, which can then be used to perform sanity checks as shown in the (self-explanatory) example below:

\begin{minted}{prolog}
max_speed(Location, S) :- reasonable_speed(Location, S1),
    posted_speed_limit(Location, S2),   
    minimum(S1, S2, S), not abnormal(Location, S).
\end{minted}  

Commonsense knowledge can also be used to ensure that the frame corresponding to a scene is consistent with information provided by various sensors in the AV. If there is any inconsistency, the sensor information can be given priority, as visual information is more likely to be erroneous. Consider the example shown in \figref{fig:tesla-av-confused}. The following rules allow \model to perform a change lane right based on sensor data overriding the visual information.
\begin{minted}{prolog}
change_lane_right_conditions(T) :-
    sensor(left, Dist, T),
    collision_distance(CD, T), Dist =< CD.
\end{minted}

Thus, ensuring that the AV system is safe is considerably easier as we are emulating a human driver's mind that gets inputs from various sources and tries to infer a consistent world view with respect to which a driving action is taken.
A system like \model can also be used to aid ML based AV systems to cross-check the decision made by them. Additionally, it's provably ethical and explainable.

\medskip\noindent{\bf Handling Complex Scenarios:}
More complex scenarios can also be handled through the use of s(CASP)-based commonsense reasoning technology. For instance, flashing traffic lights require that a sequence of scenes is processed to recognize the flashing of lights. Commonsense rules to perform this processing can be easily written in ASP/s(CASP). 
Similarly, predicting where various objects in the scene will be a few seconds in the future can also be done by analyzing a sequence of temporally ordered scene through commonsense reasoning. Incorporating more nuanced commonsense reasoning is part of our future work. 

An ML based AV system has to be retrained if it has to be used in another country with slightly different conventions. 
In contrast, a commonsense reasoning-based AV system such as \model can be easily used in such situations as the differences in conventions can be described as commonsense rules (the DL-based scene understanding system, of course, must be retrained if traffic signs are different). 
Note that humans don't have to be heavily retrained when they drive in another country. 




\section{Related Work and Conclusions}

Use of formal logic and ASP to model driving has been proposed in the past. Bhatt et al. have employed ASP and the CLINGO system for autonomous driving experiments \cite{bhatt1,bhatt2}. They propose a framework that takes visual observations computed by deep learning methods as input and provides visuo-spatial semantics at each timestamp. These semantics help in reasoning with overall scene dynamics (e.g. sudden occlusion of a motorcycle at a distance due to a car right in the front). However, their work can only \textit{support} decision-making via visual sense-making. On the other hand, our \model system focuses on ``understanding" the scene through commonsense reasoning and then computing a driving decision.
Additionally, use of CLINGO for executing ASP poses some limitations as discussed earlier \cite{arcade}. 

Karimi and Duggirala \cite{traffic-rules} have coded up rules from the California DMV handbook in ASP using CLINGO. 
Their goal is to verify the correctness of AV systems' behavior at intersections. In contrast, our approach is to use commonsense reasoning/ASP for actual autonomous driving. 
There are other works in this direction that apply
formal logic/reasoning to verifying AV systems, particularly at unsignaled intersections \cite{r1,r2,r3,r4}, as well as
situations where an AV should hand over to a human driver \cite{mccall19}.

To conclude, in this paper we described our \model system that employs commonsense reasoning (CSR) for autonomous driving. Commonsense knowledge about driving is represented as a predicate answer set program and executed on the s(CASP) goal-directed ASP system. While ML based AV have made significant advances, none of them have reached the level of full automation. We strongly believe that taking an approach based on automating CSR is indispensable for developing AV technology. The goal is to emulate human drivers, who use CSR for making decision while driving. 
Rules for driving were developed into the \model system with the help of existing data-sets such as \textsc{Kitti}, driving manuals available, and our own commonsense knowledge of driving. Our system is explainable and provably ethical. Input to the system consists of data from all the sensors as well as description of the scene as predicates (obtained using ML technology). Given that CSR is used, ML errors in processing the scene can be compensated for. 

The main contribution of our work is to demonstrate how a complete decision making system for autonomous driving can be developed by modeling commonsense reasoning in ASP and the s(CASP) system. The s(CASP) system is what makes \model possible, and it is one reason why an AV system based entirely on commonsense reasoning has not been developed thus far. 
Future work includes refining and developing the \model infrastructure to make it work with the CARLA setup (\url{https://carla.org}) as well as to develop an actual AV deployment with our industrial partner. 

\section*{Acknowledgement}

Authors are supported by NSF awards IIS 1718945, IIS 1910131, IIP 1916206 and by grants from Amazon and DoD.


%
%
%

\bibliography{bibliography_new}

\end{document}